\documentclass{article}
\usepackage{amsmath}    % higher mathematics
\usepackage{amsfonts}   % more math fonts
\usepackage{amssymb}    % more symbols
\usepackage{array}    % extensions to tabular and array
\usepackage{tabularx}   % support correct-width tables
\usepackage{times}              % typeset in Times Roman
\usepackage[numbers]{natbib}
\usepackage{epsfig}
\usepackage{multirow}
\usepackage{subfigure}
\usepackage{algorithm}
\usepackage{algorithmic}
\usepackage{booktabs}

\newcommand{\argmax}[1]{\operatornamewithlimits{arg\,max}_{#1}}
\newcommand{\argmin}[1]{\operatornamewithlimits{arg\,min}_{#1}}
\newcommand{\argsort}{\operatornamewithlimits{arg\,sort}}

\def\x{{\mathbf x}}
\def\y{{\mathbf y}}
\def\r{{\mathbf r}}

\def\q{{\mathbf q}}
\def\d{{\mathbf d}}
\def\D{{\mathbf D}}

\newcommand{\Exp}{{\sf{E\mspace{-8.7mu} E}}}  % Expectation E

\begin{document}
\title{
Loss-sensitive Training of Probabilistic Conditional Random Fields
}

\author{
\begin{tabular}{c}
    {\bf Maksims N. Volkovs} \\  
    Department of Computer Science\\ 
    University of Toronto\\
    Toronto, Canada
  \end{tabular}
  \begin{tabular}{c}
    {\bf Hugo Larochelle}\\
    D\'epartement d'informatique\\ 
    Universit\'e de Sherbrooke\\ 
    Sherbrooke, Canada
  \end{tabular}
  \begin{tabular}{c}
  {\bf Richard S. Zemel}\\  
  Department of Computer Science\\ 
  University of Toronto\\
  Toronto, Canada
  \end{tabular}
}
\date{}

\maketitle

\begin{abstract}
  We consider the problem of training probabilistic conditional random
  fields (CRFs) in the context of a task where performance is measured
  using a specific loss function.  While maximum likelihood is the
  most common approach to training CRFs, it ignores the inherent
  structure of the task's loss function. We describe alternatives to
  maximum likelihood which take that loss into account. These include
  a novel adaptation of a loss upper bound from the structured SVMs
  literature to the CRF context, as well as a new loss-inspired KL
  divergence objective which relies on the probabilistic nature of
  CRFs. These loss-sensitive objectives are compared to maximum
  likelihood using ranking as a benchmark task.  This comparison
  confirms the importance of incorporating loss information in the
  probabilistic training of CRFs, with the loss-inspired KL
  outperforming all other objectives.
\end{abstract}

\section{Introduction}

Conditional random fields (CRFs)~\cite{LaffertyJ2001} form a flexible
family of models for capturing the interaction between an input $\x$
and a target $\y$. CRFs have been designed for a vast variety of
problems, including natural language
processing~\cite{ShaF2003,SarawagiS2005,RothD2005}, speech
processing~\cite{GunawardanaA2005}, computer
vision~\cite{HeX2004,KumarS2003,QuattoniA2005} and
bioinformatics~\cite{SatoK2005,LiuY2006} tasks. One reason for their
popularity is that they provide a flexible framework for modeling the
conditional distributions of targets constrained by some specific
structure, such as chains~\cite{LaffertyJ2001},
trees~\cite{CohnT2005semanticrole}, 2D
grids~\cite{KumarS2003,NieZ2005}, permutations~\cite{VolkovsM2009} and
many more.

While there has been a lot of work on developing appropriate CRF
potentials and energy functions, as well as on deriving efficient
(approximate) inference procedures for some given target structure,
much less attention has been paid to the loss function under which the
CRF's performance is ultimately evaluated. Indeed, CRFs are usually
trained by maximum likelihood (ML) or the maximum a posteriori
criterion (MAP or regularized ML), which ignores the task's loss
function. Yet, several tasks are associated with loss
functions\footnote{Without loss of generality, for tasks where a
  performance measure is instead provided (i.e.\ where higher values
  is better), we assume it can be converted into a loss, e.g.\ by
  setting the loss to the negative of the performance measure.}  that
are also structured and do not correspond to a simple 0/1 loss:
labelwise error (Hamming loss) for item labeling, BLEU score for
machine translation, normalized discounted cumulative gain (NDCG) for
ranking, etc. Ignoring this structure can prove as detrimental to
performance as ignoring the target's structure.

The inclusion of loss information into learning is an idea that has
been more widely explored in the context of structured support vector
machines (SSVMs)~\cite{TaskarB2004,TsochantaridisI2005}. SSVMs and
CRFs are closely related models, both trying to shape an energy
or score function over the joint input and target space to fit the
available training data. However, while an SSVM attempts to satisfy
margin constraints without invoking a probabilistic interpretation of
the model, a CRF follows a probabilistic approach and instead
aims at calibrating its probability estimates to the data.
Similarly, while an SSVMs relies on maximization procedures to identify the
most violated margin constraints, a CRF relies on (approximate) inference
or sampling procedures to estimate probabilities under its
distribution and compare it to the empirical distribution.

While there are no obvious reasons to prefer one approach to the
other, a currently unanswered question is whether the known methods
that adapt SSVM training to some given loss (i.e.,\ upper bounds based
on margin and slack scaling~\cite{TsochantaridisI2005}) can also be
applied to the probabilistic training of CRFs. Another question is how
such methods would compare to other loss-sensitive training objectives
which rely on the probabilistic nature of CRFs and which may have no
analog in the SSVM framework.

We investigate these questions in this paper. First, we
describe upper bounds similar to the margin and slack scaling upper
bounds of SSVMs, but that correspond to maximum likelihood training
of CRFs with loss-augmented and loss-scaled energy
functions. Second, we describe two other loss-inspired training
objectives for CRFs which rely on the probabilistic nature of CRFs:
the standard average expected loss objective and a novel loss-inspired
KL-divergence objective. Finally, we compare these loss-sensitive objectives 
on ranking benchmarks based on the NDCG performance
measure. To our knowledge, this is the first systematic evaluation of
loss-sensitive training objectives for probabilistic CRFs.

\section{Conditional Random Fields}

This work is concerned with the general problem of supervised learning, where
the relationship between an input $\x$ and a target $\y$ must be learned from a
training set of instantiated pairs ${\cal D} = \{\x_t,\y_t\}$. More
specifically, we are interested in learning a predictive mapping from $\x$ to
$\y$. 

Conditional random fields (CRFs) tackle this problem by defining
directly the conditional distribution $p(\y|\x)$ through
some energy function $E(\y,\x;\theta)$ as follows:
\begin{equation}
  p(\y|\x) = \exp(-E(\y,\x;\theta)) / Z(\x),~~Z(\x) = 
		\sum_{\y \in {\cal Y}(\x)} \exp(-E(\y,\x;\theta))
\end{equation}
where ${\cal Y}(\x)$ is the set of all possible configurations for
$\y$ given the input $\x$ and $\theta$ is the model's parameter
vector. The parametric form of the energy function $E(\y,\x;\theta)$
will depend on the nature of $\x$ and $\y$. A popular choice is that
of a linear function of a set of features on $\x$ and $\y$,
i.e.,\ $E(\y,\x;\theta) = - \sum_i \theta_i f_i(\x,\y)$.

\subsection{Maximum Likelihood Objective}

The most popular approach to training CRFs is conditional maximum
likelihood. 
It corresponds to the minimization with respect to $\theta$ 
of the objective ${\cal L}_{\rm ML}({\cal D};\theta)$:
\begin{equation}
  - \frac{1}{|{\cal D}|} \sum_{(\x_t,\y_t) \in {\cal D}} \log p(\y_t|\x_t) =
\frac{1}{|{\cal D}|} \sum_{(\x_t,\y_t) \in {\cal D}}  E(\y_t,\x_t;\theta)
+\log\left( \sum_{\y \in {\cal Y}(\x)} \exp(-E(\y,\x_t;\theta))\right)~.
\label{eqn:ml}
\end{equation}
To this end, one can use any gradient-based optimization procedure,
which can convergence to a local optimum, or even a global optimum if
the problem is convex (e.g.,\ by choosing an energy function
$E(\y,\x;\theta)$ linear in $\theta$). The gradients have an elegant
form:
\begin{equation}
  \frac{\partial {\cal L}_{\rm ML}({\cal D};\theta)}{\partial \theta} = 
	\frac{1}{|{\cal D}|} \sum_{(\x_t,\y_t) \in {\cal D}}
	\frac{\partial E(\y_t,\x_t;\theta)}{\partial \theta} -
	\Exp_{\y|\x_t}\left[\frac{\partial
	E(\y,\x_t;\theta)}{\partial \theta}\right]. \label{eqn:grad}
\end{equation}
Hence exact gradient evaluations are possible when the conditional
expectation in the second term is tractable. This is the case for
CRFs with a chain or tree structure, for which belief propagation
can be used. When gradients are intractable, two approximate alternatives can be
considered. The first is to approximate the intractable expectation
using either Markov chain Monte Carlo sampling or variational
inference algorithms such as mean-field or loopy belief propagation,
the latter being the most popular. The second approach is to use
alternative objectives such as pseudolikelihood~\cite{BesagJ1975} or
piece-wise training~\cite{SuttonC2005}\footnote{Variational
inference-based training can also be interpreted as training based
on a different objective.}.

\section{Loss-sensitive Training Objectives}
\label{sect:loss-objectives}

Unfortunately, maximum likelihood and its associated approximations
all suffer from the problem that the loss function under which the
performance of the CRF is evaluated is ignored. In the well-specified
case and for large datasets, this would probably not be a problem
because of the asymptotic consistency and efficiency properties of
maximum likelihood. However, almost all practical problems do not fall
in the well-specified setting, which justifies the exploration of
alternative training objectives.

Let $\hat{\y}(\x_t)$ denote the prediction made by a CRF for some given
input $\x_t$. Most commonly\footnote{For loss functions that decompose
  into loss terms over subsets of target variables, it may be more
  appropriate to use the mode of the marginals over each subset as the
  prediction.}, this prediction will be $\hat{\y}(\x_t) = \argmax{\y
  \in {\cal Y}(\x_t)} p(\y|\x_t) = \argmin{\y \in {\cal Y}(\x_t)}
E(\y,\x_t)$.  We assume that we are given some loss
$l_t(\hat{\y}(\x_t))$ under which the performance of the CRF on some
dataset ${\cal D}$ will be measured. We will also assume that
$l_t(\y_t) = 0$. The goal is then to achieve a low average
$\frac{1}{|{\cal D}|} \sum_{(\x_t,\y_t) \in {\cal D}}
l_t(\hat{\y}(\x_t))$ under that loss. 

Directly minimizing this average loss is hard, because
$l_t(\hat{\y}(\x_t))$ is not a smooth function of the CRF parameters
$\theta$. In fact, the loss itself $l_t(\hat{\y}(\x_t))$ is normally
not a smooth function of the prediction $\hat{\y}(\x_t)$, and
$\hat{\y}(\x)$ is also not a smooth function of the model parameters
$\theta$. This non-smoothness makes it impossible to apply
gradient-based optimization.

However, one could attempt to indirectly optimize the average loss by
deriving smooth objectives that also depend on the loss. In the next
sections, we describe three separate formulations of this approach.

\subsection{Loss Upper Bounds}
\label{sect:lossbounds}

The loss function provides important information as to how good a
potential prediction $\y$ is with respect to the ground truth
$\y_t$. In particular, it specifies an ordering from the best
prediction ($\y = \y_t$) to increasingly bad predictions with
increasing value of their associated loss $l_t(\y)$. It might then be
desirable to ensure that the CRF assigns particularly low probability
(i.e.,\ high energy) to the worst possible predictions, as measured by
the loss.

A first way of achieving this is to augment the energy function
at a given training example $(\x_t,\y_t)$ by including 
the loss function for that example, producing a {\it Loss-Augmented} energy:
\begin{equation}
  E^{\rm LA}_t(\y,\x_t;\theta) = E(\y,\x_t;\theta) - l_t(\y)~.
\end{equation}
By artificially reducing the energy of bad values of $\y$ as a
function of their loss, this will force the CRF to increase even more
the value of $E(\y,\x;\theta)$ for those values of $\y$ with high
loss. This idea is similar to the concept of margin re-scaling in
structured support vector machines
(SSVMs)~\cite{TaskarB2004,TsochantaridisI2005}, a similarity that has
been highlighted previously by \citet{HazanT2010}.  Moreover, as in
SSVMs, it can be shown that by replacing the regular energy function
with this loss-augmented energy function in the maximum likelihood
objective of Equation~\ref{eqn:ml}, we obtain a new
Loss-Augmented objective that upper bounds the average loss:
\begin{eqnarray*}
  {\cal L}_{\rm LA}({\cal D};\theta)& = & \frac{1}{|{\cal D}|} 
\sum_{(\x_t,\y_t) \in {\cal D}} E^{\rm LA}_t(\y_t,\x_t) + \log\left( \sum_{\y
\in {\cal Y}(\x)} \exp(-E^{\rm LA}_t(\y,\x_t))\right)\\
  &\geq& \frac{1}{|{\cal D}|}\sum_{(\x_t,\y_t) \in {\cal D}} 
E^{\rm LA}_t(\y_t,\x_t) + \log \left( \exp(-E^{\rm
LA}_t(\hat{\y}(\x_t),\x_t))\right) \\
  &=&\frac{1}{|{\cal D}|}\sum_{(\x_t,\y_t) \in {\cal D}} E(\y_t,\x_t) -
E(\hat{\y}(\x_t),\x_t) + l_t(\hat{\y}(\x_t)) - l_t(\y_t)\\
  &\geq& \frac{1}{|{\cal D}|}\sum_{(\x_t,\y_t) \in {\cal D}}l_t(\hat{\y}(\x_t)) 
~.
\end{eqnarray*}
We see that the higher $l_t(\y)$ is for some given $\y$, the more
important the energy term associated with it will be in the global
objective. Hence, introducing the loss this way will indeed force the
optimization to focus more on increasing the energy for configurations
of $\y$ associated with high loss.

As an alternative to subtracting the loss, we could further increase
the weight of terms associated with high loss by also multiplying the
original energy function, as follows:
\begin{equation} \label{eq:loss_scaled}
E^{\rm LS}_t(\y,\x_t;\theta) = l_t(\y) ( E(\y,\x_t;\theta) - 
	E(\y_t,\x_t;\theta) ) - l_t(\y) ~.
\end{equation}
The advantage of this {\it Loss-Scaled} energy is that when a configuration
of $\y$ with high loss already has higher energy than the target
(i.e.,\ $E(\y,\x_t;\theta) - E(\y_t,\x_t;\theta) > 0$), then the energy
is going to be further increased, reducing its weight in the
optimization.  In other words, focus in the optimization is put on
bad configurations of $\y$ only when they have lower energy than the
target. Finally, we can also show that the Loss-Scaled objective
obtained from this loss-scaled energy leads to an upper bound on the average
loss: 
\begin{eqnarray*}
  {\cal L}_{\rm LS}({\cal D};\theta)&=& \frac{1}{|{\cal D}|} 
\sum_{(\x_t,\y_t) \in {\cal D}} E^{\rm LS}_t(\y_t,\x_t) + \log\left( \sum_{\y
\in {\cal Y}(\x)} \exp(-E^{\rm LS}_t(\y,\x_t))\right)\\
  &=& \frac{1}{|{\cal D}|} \sum_{(\x_t,\y_t) \in {\cal D}} \log\left( \sum_{\y 
\in {\cal Y}(\x)} \exp(- l_t(\y) ( E(\y,\x_t;\theta) - E(\y_t,\x_t;\theta) ) +
l_t(\y))\right)\\
  & \geq & \frac{1}{|{\cal D}|}\sum_{(\x_t,\y_t) \in {\cal D}} l_t(\hat{\y}
(\x_t))( 1 + E(\y_t,\x_t;\theta) - E(\hat{\y}(\x_t),\x_t;\theta) ) \\
& \geq & \frac{1}{|{\cal D}|}\sum_{(\x_t,\y_t) \in {\cal D}}l_t(\hat{\y} 
(\x_t))~.
\end{eqnarray*}
There is a connection with SSVM training objectives here as well. This
loss-scaled CRF is the probabilistic equivalent of SSVM training with
slack re-scaling~\cite{TsochantaridisI2005}.

Since both the loss-augmented and loss-scaled CRF objectives follow
the general form of the maximum likelihood objective but with
different energy functions, the form of the gradient is also that of
Equation~\ref{eqn:grad}. The two key differences are that the energy
function is now different, and the conditional expectation on $\y$
given $\x_t$ is according to the CRF distribution with the associated
loss-sensitive energy. In general (particularly for the loss-scaled
CRF), it will not be possible to run belief propagation to compute the
expectation\footnote{In the loss-augmented case, one exception is if
  the loss decomposes into individual losses over each target variable
  $y_i$ and the CRF follows a tree structure in its output. In this
  case, the loss terms can be integrated into the CRF unary features
  and belief propagation will perform exact inference.}, but adapted
forms of loopy belief propagation or MCMC (e.g.,\ Gibbs sampling) could
be used.

\subsection{Expected Loss}
\label{sect:exploss}

A second approach to deriving a smooth version of the average loss is
to optimize the average {\it Expected Loss}, where the expectation is based
on the CRF's distribution:
\begin{equation}
{\cal L}_{\rm EL}({\cal D};\theta) = \frac{1}{|{\cal D}|} 
	\sum_{(\x_t,\y_t) \in {\cal D}} \Exp_{\y|\x_t}\left[ l_t(\y)\right] =
	\frac{1}{|{\cal D}|} \sum_{(\x_t,\y_t) \in {\cal D}} \sum_{\y \in {\cal
	Y}(\x_t)} l_t(\y) p(\y|\x_t)~.
\end{equation}
While this objective is not an upper bound, it becomes increasingly
closer to the average loss as the entropy of $p(\y|\x_t)$ becomes
smaller and puts all its mass on $\hat{\y}(\x_t)$.

The parameter gradient has the following form:
\begin{equation}
\frac{\partial {\cal L}_{\rm EL}({\cal D};\theta)}{\partial \theta}  =  
	\frac{1}{|{\cal D}|} \sum_{(\x_t,\y_t) \in {\cal D}}
	\Exp_{\y|\x_t}\left[l_t(\y)\right] \Exp_{\y|\x_t}\left[\frac{\partial
	E(\y,\x_t;\theta)}{\partial \theta}\right] - \Exp_{\y|\x_t}\left[
	l_t(\y) \frac{\partial E(\y_t,\x_t;\theta)}{\partial \theta}\right] ~.
\end{equation}
If the required expectations cannot be computed tractably, MCMC
sampling can be used to approximate them. Another alternative
is to use a fixed set of representative samples \cite{VolkovsM2009}.

\subsection{Loss-inspired Kullback-Leibler}

Both the average expected loss and the loss upper bound objectives
have in common that their objectives are perfectly minimized when the
posteriors $p(\y|\x_t)$ put all their mass on the targets $\y_t$. In practice,
this is bound not to happen, since this is likely to correspond
to an overfitted solution which will be avoided using additional regularization.

Instead of relying on a generic regularizer such as the $\ell^2$-norm
of the parameter vector, perhaps the loss function itself might
provide cues as to how best to regularize the CRF. Indeed, we can
think of the loss as a ranking of all potential predictions, from
perfect to adequate to worse. Hence, if we are not to put all 
probability mass on $p(\y_t|\x_t)$, we could make use of the information
provided by the loss in order to determine how to distribute the
excess mass $1-p(\y_t|\x_t)$ on other configurations of $\y$. In particular,
it would be sensible to distribute it on other values of $\y$ in proportion to
the loss $l_t(\y)$.

To achieve this, we propose to first convert the loss into a
distribution over the target $q(\y | t)$ and then minimize the
{\it Kullback-Leibler} (KL) divergence between this target distribution and
the CRF posterior:
\begin{equation}
{\cal L}_{\rm KL}({\cal D};\theta) = \frac{1}{|{\cal D}|} 
\sum_{(\x_t,\y_t) \in {\cal D}} D_{\rm KL}(q(\cdot|t) || p(\cdot | \x_t)) = -
\frac{1}{|{\cal D}|} \sum_{(\x_t,\y_t) \in {\cal D}} \sum_{\y \in {\cal
Y}(\x_t)} q(\y|t) \log p(\y|\x_t) - C
\end{equation}
where constant $C=H(q(\cdot|t))$ is the entropy of the target
distribution, which does not depend on parameter vector $\theta$.

\begin{figure}[t]
\centerline{
	\includegraphics[scale=0.30]{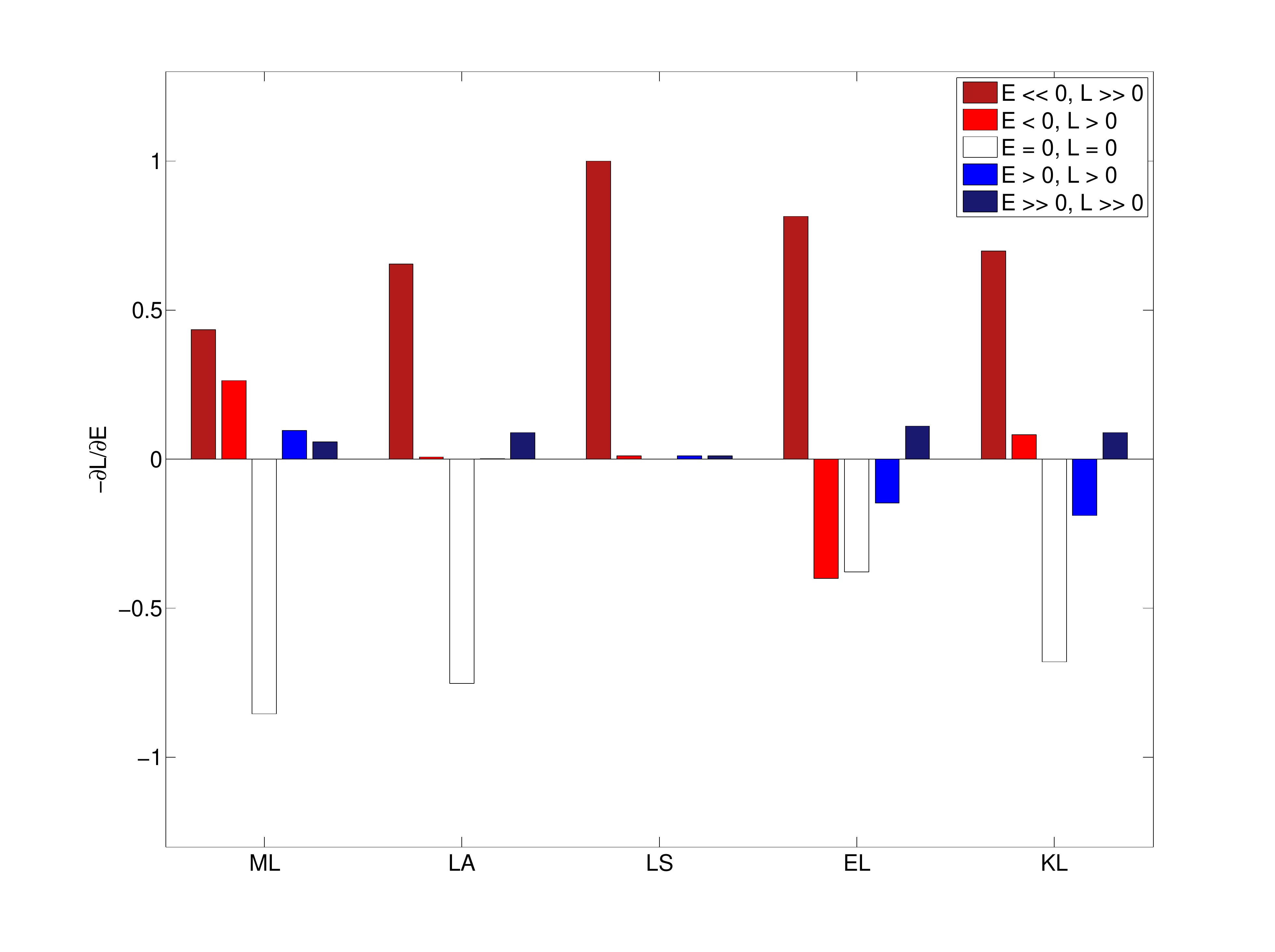}
}
\caption{Negative derivatives of the objective with respect to energy for each
of the five training objectives presented. 
The five training objectives ar: maximum likelihood (ML),
loss-augmented ML (LA), loss-scaled ML (LS), expected loss (EL) and
loss-inspired Kullback-Leibler divergence (KL). 
For each objective we consider five
different configurations: from left to right, their 
energies are $[-1, -0.5, 0, 0.5, 1]$ and the losses are $[5, 1, 0, 1, 5]$.
The middle one therefore corresponds to a ground-truth configuration;
those to its left are currently more likely under the model, and
loss increases with distance from this middle one.
The derivatives for each objective are normalized by the $\ell^2$ norm.}
\label{fig:loss_behavior}
\end{figure}

There are several ways of defining the target distribution $q(\y|t)$. In this
work, we define it as follows: 
\begin{equation}
q(\y|t) = \exp(-l_t(\y) / T) / Z_t,~~Z_t = \sum_{\y\in{\cal Y}(x_t)}  
\exp(-l_t(\y) / T) 
\end{equation}
where the temperature parameter $T$ controls how peaked this
distribution is around $\y_t$. The maximum likelihood objective
is recovered as $T$ approaches 0.

The gradient with respect to $\theta$ is simply the expectation of the
gradient for maximum likelihood ${\cal L}_{\rm ML}$ according to the
target distribution $q(\y | t)$. Here too, if the expectation is not
tractable, one can using sampling to approximate it. In particular,
since we have total control over the form of $q(\y|t)$, it is easy
to define it such that it can be sampled from exactly.

\subsection{Analyzing the Behavior of the Training Objectives}

Figure \ref{fig:loss_behavior} shows how the gradient with respect to the energy
changes for each objective as we consider configurations $\y$ with varying energy and
loss values. 
From this figure we see significant differences in the behaviors of the introduced
objective functions. Only the expected-loss and loss-inspired Kullback-Leibler objectives
will attempt to lower the energies of configurations that have non-zero loss. The maximum
likelihood objective aims to raise the energies of the non-zero loss configurations,
in proportion to how probable they are. On the other
hand the loss-augmented and loss-scaled objectives concentrate on the most probable
configurations that have the highest loss (worst violators), with the loss-scaled
objective having the most extreme behavior and putting all the gradient on the
worst violator. This behavior is expected as the energies get amplified by the
addition (multiplication) of the loss which artificially raises the probability
of the already probable violators.

The behavior of the expected-loss objective is counter-intuitive as it
tries to lower the energy of all configurations that have low loss,
including those that are already more probable than the zero-loss
one. In this example, it even pushes down more the energy of a
non-zero loss configuration more than that of the zero-loss (target)
configuration. The loss-inspired KL objective adjusts this and only lowers
the energy of the zero-loss (ground-truth) and the low-loss configuration that has
low probability.

\section{Learning with Multiple Ground Truths}
\label{sect:multigt}

In certain applications, for some given input $\x_t$, there is not
only a single target $\y_t$ that is correct (see Section~\ref{sect:exp} for the
case of ranking). This information can easily be encoded within the loss
function, by setting $l_t(\y) = 0$ for all such valid predictions.

In this context, maximum likelihood training corresponds to the objective:
\begin{equation}
  {\cal L}_{\rm ML}({\cal D};\theta) = - \frac{1}{|{\cal D}|} 
\sum_{(\x_t,\r_t) \in {\cal D}} ~\sum_{\y_t \in {\cal Y}_0(\x_t)} \log
p(\y_t|\x_t)
\end{equation}
where ${\cal Y}_0(\x_t) = \{\y | \y \in {\cal Y}(\x_t), \ l_t(\y)=0\}$. This is
equivalent to maximizing the likelihood of all predictions $\y$ that are
consistent with the loss, i.e.\ that have zero loss. The loss-augmented variant
is similarly adjusted. As for loss-scaling, we replace the energy at the ground
truth with the average energy of all valid ground truths in the loss-scaled
energy:
\begin{equation}
  E^{\rm LS}_t(\y,\x_t;\theta) = l_t(\y) \left( E(\y,\x_t;\theta) - 
\frac{1}{|{\cal Y}_t(\x_t)|} \sum_{\y_t \in {\cal Y}_0(\x_t)}
E(\y_t,\x_t;\theta) \right) - l_t(\y) ~.
\end{equation}
No changes to the average expected loss and loss-inspired KL objectives are
necessary as they consider all valid $\y$.

In the setting of multiple ground truths, a clear distinction can be made
between the average expected loss and the other objectives, in terms of the
solutions they encourage. Indeed, the expected loss will be minimized as long as
$\sum_{\y_t \in {\cal Y}_t(\x_t)} p(\y_t|\x_t) = 1$, i.e.\ probability mass is
only put on configurations of $\y$ that have zero loss. On the other hand, the
maximum likelihood and loss upper bound objectives add the requirement that the
mass be equally distributed amongst those configurations. As for the
loss-inspired KL, it requires that the sum of the probability mass sum to a
constant smaller than 1, specifically $1- \sum_{\y_t \in {\cal Y}_t(\x_t)}
q(\y_t | t)$.

\section{Related Work}

While maximum likelihood is the dominant approach to training CRFs in
the literature, others have proposed ways of adapting the CRF training
objective for specific tasks. For sequence labeling problems,
\citet{KakadeS2002} proposed to maximize the label-wise marginal
likelihood instead of the joint label sequence likelihood, to reflect
the fact that the task's loss function is the sum of label-wise
classification errors. \citet{SuzukiJ2006,GrossS2007} went a step
further by proposing to directly optimize a smoothed version of the
label-wise classification error (\citet{SuzukiJ2006} also described
how to apply it to optimize an F-score). Their approach is similar to
the average expected loss described in Section~\ref{sect:exploss},
however they do not discuss how to generalize it to arbitrary loss
functions. The average expected loss objective for CRFs was formulated
by \citet{TaylorM2008} and \citet{VolkovsM2009}, in the context of ranking.

Work in other frameworks than CRFs for structured output prediction
have looked at how to incorporate loss information into learning.
\citet{TsochantaridisI2005} describe how to upper bound the average
loss with margin and slack scaling.  \citet{McAllesterD2010} propose a
perceptron-like algorithm based on an update which in expectation is
close to the gradient on the true expected loss (i.e.,\ the
expectation is with respect to the true generative process). Both
SSVMs and perceptron algorithms require procedures for computing the
so-called loss-adjusted MAP assignment of the output $\y$ which, for
richly structured losses, can be intractable. One advantage of CRFs is
that they can instead leverage the vast MCMC literature to sample from
CRFs with loss-adjusted energies. Moreover, they open the door to
alternative (i.e. not necessarily upper-bounding) objectives.

Finally, while \citet{HazanT2010} described how margin scaling can be applied to
CRFs, we give for the first time the equivalent of slack scaling
for CRFs in Section~\ref{sect:lossbounds}.

\section{Experiments}
\label{sect:exp}

We evaluate the usefulness of the different loss-sensitive training
objectives on ranking tasks. In this setting, the input $\x=\left(\q,\D\right)$
corresponds to a pair made of a query vector $\q$ and a set of documents
$\D=\{\d^{(i)}\}$, and $\y$ is a vector corresponding to a ranking\footnote{For
example, if $y_{i} = 3$, then document $\d^{(i)}$ is ranked third amongst all
documents $\D$ for the query $\q$.} of each document $\d^{(i)}$ among the whole
set of documents $\D$.

Ranking is particularly interesting as a benchmark task for loss-sensitive
training of CRFs for two reasons. The first is the complexity of the output
space ${\cal Y}(\q,\D)$, which corresponds to all possible permutations of
documents $\D$, making the application of CRFs to this setting more challenging
than sequential labeling problems with chain structure.

The second is that learning to rank is an example of a task with
multiple ground truths (see Section~\ref{sect:multigt}), which is a
more challenging setting than the single ground truth case. Indeed,
for each training input $\x_t = (\q_t,\D_t)$, we are not given a
single target rank $\y_t$, but a vector $\r_t$ of relevance level
values for each document. The higher the level, the more relevant the
document is and the better its rank should be. Moreover, two documents
$\d_t^{(i)}$ and $\d_t^{(j)}$ with the same relevance level (i.e.,\
$r_{ti} = r_{tj}$) are indistinguishable in their ranking, meaning
that they can be swapped within some ranking without affecting the
quality of that ranking.

The quality of a ranking is measured by the Normalized Discounted
Cumulative Gain:
\begin{equation}
  NDCG(\y,\r_t) = N_t \sum_{i=1}^{m_t} \frac{r_{ti} \log(2)}{\log(1 +
y_i)} 
\end{equation}
where $N_t = 1 / NDCG(\argsort(-\r_t),\r_t)$ is a normalization constant that
insures the maximum value of NDCG is 1, which is achieved when documents are
ordered in decreasing order of their relevance levels. Note that this is not a
standard definition of NDCG, we use it here because this form was adopted for
evaluation of the baselines on the Microsoft's LETOR4.0 datset collection
\cite{LETOR}. To convert NDCG into a loss, we simply define $l_t(\y) =
1-NDCG(\y,\r_t)$.
\begin{figure}[t]
\centerline{
	\subfigure[MQ2007]{
		\includegraphics[scale=0.45]{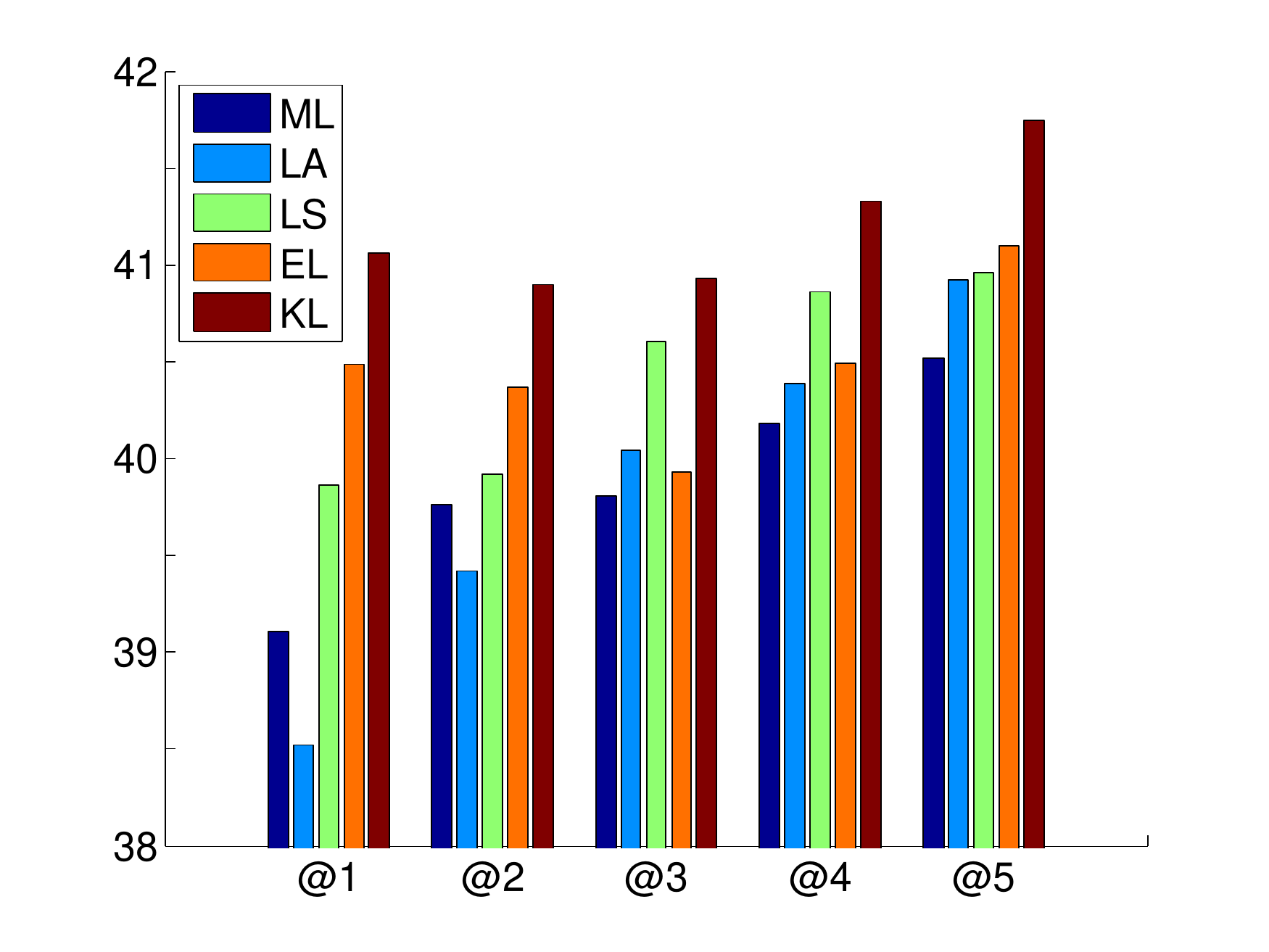}
		\label{fig:MQ2007results}
	}
	\subfigure[MQ2008]{
		\includegraphics[scale=0.45]{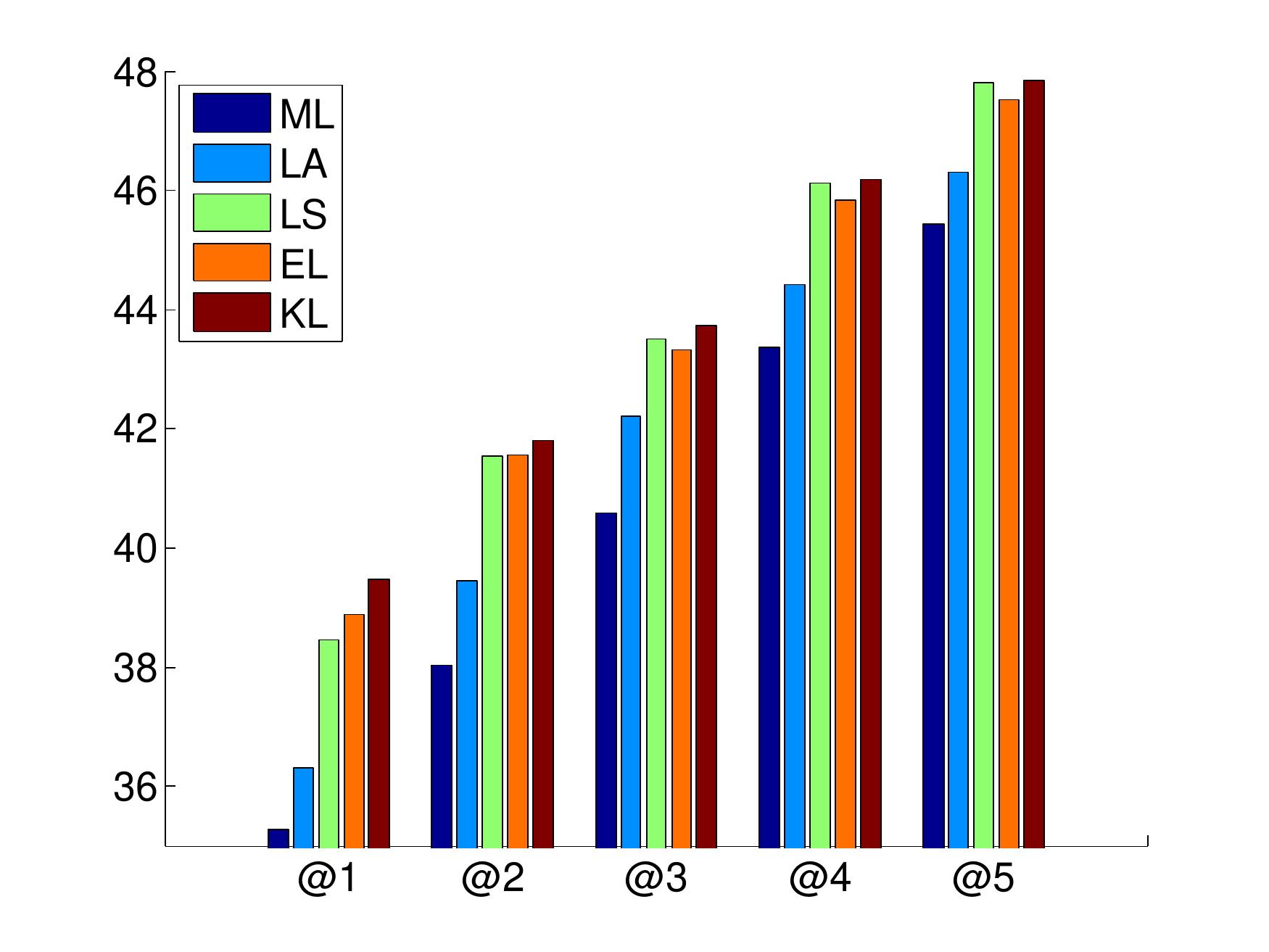}
		\label{fig:MQ2008results}
	}
}
\caption{NDCG@1-5 results on MQ2008 and MQ2007 datasets for different learning
objectives.}

\end{figure}

A common approach to ranking is to learn a scoring function $f(\q,\d^{(i)})$
which outputs for all documents $\d^{(i)}\in\D$ a score corresponding to
how relevant document $\d^{(i)}$ is for query $\q$. Here,
we follow the same approach by incorporating this scoring function
into the energy function of the CRF. We use an energy function
linear in the scores:
\begin{equation}\label{eqn:unary-energy}
  E(\y,\q,\D) = \sum_{i=1}^{|\D|} \alpha_{y_i} f(\q,\d^{(i)})
\end{equation}
where $\alpha$ is a weight vector of decreasing values (i.e.,\
$\alpha_i > \alpha_j$ for $i<j$). In our experiments, we use a
weighting inspired by the NDCG measure: $\alpha_i = \log(2) /
\log(i+1)$.  Using this energy function, we can show that the
prediction $\hat{\y}(\q,\D)$ is obtained by sorting the documents in
decreasing order of their scores:
\begin{equation}
\hat{\y}(\q,\D) = \argmin{\y \in {\cal Y}(\q,\D)} E(\y,\q,\D) = 
		\argsort([-f(\q,\d^{(1)}),\dots,-f(\q,\d^{(|\D|)})])~.
\end{equation}
As for the scoring function, we use a simple linear function
$f(\q,\d^{(i)}) = \theta^T \phi(\q,\d^{(i)})$ on a joint
query-document feature representation $\phi(\q,\d^{(i)})$. A standard
feature representation is provided in each ranking datasets we
considered.

We trained CRFs according to maximum likelihood as well as the different
loss-sensitive objectives described in Section~\ref{sect:loss-objectives}.  In
all cases, stochastic gradient descent was used by iterating over queries and
performing a gradient step update for each query. Moreover, because the size of
${\cal Y}(\q,\D)$ is factorial in the number of documents, explicit summation
over that set is only tractable for a small number of documents. To avoid this
problem we use an approach similar to the one suggested by Petterson et al.
\cite{Vis_GraphMatching}. Every time a query $q_t$ is visited and its
associated set of documents $\D_t$ is greater than $6$, we randomly select a
subset of $6$ documents $\widetilde{\D} \subset \D_t$, ensuring that it contains at
least one document of every relevance level found for that query. The exact
parameter gradients can then be computed for this reduced set by enumerating
all possible permutations, and the CRF can be updated.

\subsection{Datasets}

In our experiments we use the LETOR \cite{LETOR} benchmark datasets. These data
sets were chosen because they are publicly available, include several baseline
results, and provide evaluation tools to ensure accurate comparison between
methods. In LETOR4.0 there are two learning to rank data sets MQ2007 and MQ2008.
MQ2007 contains 1692 queries with 69623 documents and MQ2008 contains 784
queries and a total of 15211 documents. Each query document pair is assigned one
of three relevance judgments: 2 = highly relevant, 1 = relevant and
0 = irrelevant. Both datasets come with five precomputed folds with
60/20/20 slits for training validation and testing. The results show for each
model the averages of the test set results for the five folds.

\subsection{Results}

We experimented with five objective functions, namely: maximum likelihood (ML),
loss-augmented ML (LA), loss-scaled ML (LS), expected loss (EL) and
loss-inspired Kullback-Leibler divergence (KL). For the loss-augmented objective 
we introduced an additional weight
$\alpha > 0$ modifying the energy to: $E_t(\y, \x_t; \theta) = E(\y, \x_t;
\theta) - \alpha l_t(\y)$. In this form $\alpha$ controls the contribution of
the loss to the overall energy. For all objectives we did a sweep over
learning rates in [0.5, 0.01, 0.01, 0.001]. Moreover we experimented 
with $\alpha$ in [1, 10, 20, 50] for the loss-augmented objective and $T$ in
[1, 10, 20, 50] for the KL objective. For each fold the setting that gave the
best validation NDCG was chosen and the corresponding model was then tested on
the test set.
\begin{table}[t]\centering
\caption{NDCG@1-5 results on MQ2008 and MQ2007 datasets.}
\begin{tabular}{rccccccccccc}
\toprule
\phantom{ab} & \multicolumn{5}{c}{MQ2007: NDCG} & \phantom{ab} &
\multicolumn{5}{c}{MQ2008: NDCG}\\
\cmidrule{2-6} \cmidrule{8-12}
& @1 & @2 & @3 & @4 & @5 && @1 & @2 & @3 & @4 & @5\\
\midrule
Regression& 38.94& 39.60& 39.86& 40.53& 41.11&&
		36.67& 40.62& 42.89& 45.60& 47.73\\
SVM-Struct& 40.96 &40.73 &40.62 &40.84 &41.42&& 
		36.26& 39.84& 42.85& 45.08& 46.95\\
ListNet& 40.02 &40.63	&40.91 &\bf{41.44} &41.70&&
		37.54& 41.12& 43.24& 45.68& 47.47\\
AdaRank& 38.76& 39.67& 40.44& 40.67& 41.02&&
		38.26& \bf{42.11}& \bf{44.20}& \bf{46.53}& \bf{48.21}\\
KL& \bf{41.06}& \bf{40.90}& \bf{40.93}& 41.33& \bf{41.75}&&
		\bf{39.47}& 41.80& 43.74& 46.18& 47.84\\
\bottomrule
\end{tabular}
\label{tb:baselines}
\end{table}

The results for the five objective functions are shown in Figures
\ref{fig:MQ2007results} and \ref{fig:MQ2008results}. First, we see
that in almost all cases loss-augmentation produces better results than
the base maximum likelihood approach. Second, loss-scaling further improves
on the loss-augmentation results and has similar performance to the expected
objective. Finally, among all objectives, KL consistently produces the best
results on both datasets. Taken together, these results strongly support our claim that
incorporating the loss into the learning procedure of CRFs is important.

Comparisons of the CRFs trained on the KL objective with other models
is also shown in Table \ref{tb:baselines}, where the performance of linear
regression and other linear baselines listed on LETOR's website is provided. 
We see that KL outperforms the baselines on the MQ2007 dataset on all
truncations except $4$. Moreover, on MQ2008 the performance KL is comparable to
the best baseline AdaRank, with KL beating AdaRank on NDCG@1. We note also that
KL consistently outperforms LETOR's SVM-Struct baseline.

\section{Conclusion}

In this work, we explored different approaches to incorporating loss
function information into the training objective of a probabilistic
CRF. We discussed how to adapt ideas from the SSVM literature to
the probabilistic context of CRFs, introducing for the first time 
the equivalent of slack scaling to CRFs. We also described objectives
that rely on the probabilistic nature of CRFs, including a novel
loss-inspired KL objective. In an empirical comparison on ranking
benchmarks, this new KL objective was shown to consistently outperform
all other loss-sensitive objectives. 

To our knowledge, this is the broadest comparison of loss-sensitive
training objectives for probabilistic CRFs yet to be made. It
strongly suggests that the most popular approach to CRF training,
maximum likelihood, is likely to be suboptimal. While ranking was
considered as the benchmark task here, in future work, we would like
to extend our empirical analysis to other tasks such as labeling tasks.

\bibliographystyle{unsrtnat}
\bibliography{crf_training_objectives}

\begin{thebibliography}{25}
\providecommand{\natexlab}[1]{#1}
\providecommand{\url}[1]{\texttt{#1}}
\expandafter\ifx\csname urlstyle\endcsname\relax
  \providecommand{\doi}[1]{doi: #1}\else
  \providecommand{\doi}{doi: \begingroup \urlstyle{rm}\Url}\fi

\bibitem[Lafferty et~al.(2001)Lafferty, McCallum, and Pereira]{LaffertyJ2001}
John Lafferty, Andrew McCallum, and Fernando Pereira.
\newblock Conditional random fields: {P}robabilistic models for segmenting and
  labeling sequence data.
\newblock In \emph{ICML}, pages 282--289. Morgan Kaufmann, San Francisco, CA,
  2001.

\bibitem[Sha and Pereira(2003)]{ShaF2003}
Fei Sha and Fernando Pereira.
\newblock Shallow parsing with conditional random fields.
\newblock In \emph{HLT/NAACL}, 2003.

\bibitem[{Sarawagi} and {Cohen}(2005)]{SarawagiS2005}
Sunita {Sarawagi} and William~W. {Cohen}.
\newblock Semi-markov conditional random fields for information extraction.
\newblock In Lawrence~K. Saul, Yair Weiss, and {L\'{e}on} Bottou, editors,
  \emph{NIPS}, pages 1185--1192. MIT Press, Cambridge, MA, 2005.

\bibitem[Roth and Yih(2005)]{RothD2005}
Dan Roth and Wen-tau Yih.
\newblock Integer linear programming inference for conditional random fields.
\newblock In \emph{Proceedings of the 22nd international conference on Machine
  learning}, ICML, pages 736--743, New York, NY, USA, 2005. ACM.

\bibitem[Gunawardana et~al.(2005)Gunawardana, Mahajan, Acero, and
  Platt]{GunawardanaA2005}
Asela Gunawardana, Milind Mahajan, Alex Acero, and John~C. Platt.
\newblock Hidden conditional random fields for phone classification.
\newblock In \emph{Interspeech}, pages 1117--1120, 2005.

\bibitem[He et~al.(2004)He, Zemel, and Carreira-Perpi{\~n}{\'a}n]{HeX2004}
Xuming He, Richard~S. Zemel, and Miguel~{\'A}. Carreira-Perpi{\~n}{\'a}n.
\newblock Multiscale conditional random fields for image labeling.
\newblock In \emph{CVPR}, pages 695--702, 2004.

\bibitem[Kumar and Hebert(2003)]{KumarS2003}
Sanjiv Kumar and Martial Hebert.
\newblock Discriminative fields for modeling spatial dependencies in natural
  images.
\newblock In \emph{NIPS}, 2003.

\bibitem[Quattoni et~al.(2005)Quattoni, Collins, , and Darrell]{QuattoniA2005}
Ariadna Quattoni, Michael Collins, , and Trevor Darrell.
\newblock Conditional random fields for object recognition.
\newblock In \emph{NIPS}, 2005.

\bibitem[Sato and Sakakibara(2005)]{SatoK2005}
Kengo Sato and Yasubumi Sakakibara.
\newblock Rna secondary structural alignment with conditional random fields.
\newblock \emph{Bioinformatics}, 2005.

\bibitem[Liu et~al.(2006)Liu, Carbonell, Weigele, and Gopalakrishnan]{LiuY2006}
Yan Liu, Jaime Carbonell, Peter Weigele, and Vanathi Gopalakrishnan.
\newblock Protein fold recognition using segmentation conditional random fields
  (scrfs).
\newblock \emph{Journal of Computational Biology}, 2006.

\bibitem[Cohn and Blunsom(2005)]{CohnT2005semanticrole}
Trevor Cohn and Philip Blunsom.
\newblock Semantic role labelling with tree conditional random fields.
\newblock In \emph{CoNLL}, pages 169--172, 2005.

\bibitem[Nie et~al.(2005)Nie, rong Wen, Zhang, and ying Ma]{NieZ2005}
Zaiqing Nie, Ji~rong Wen, Bo~Zhang, and Wei ying Ma.
\newblock 2d conditional random fields for web information extraction.
\newblock In \emph{ICML}, pages 1044--1051. ACM Press, 2005.

\bibitem[Volkovs and Zemel(2009)]{VolkovsM2009}
Maksims Volkovs and Richard Zemel.
\newblock {BoltzRank}: Learning to maximize expected ranking gain.
\newblock In L\'{e}on Bottou and Michael Littman, editors, \emph{ICML}, pages
  1089--1096, Montreal, June 2009. Omnipress.

\bibitem[Taskar et~al.(2004)Taskar, Guestrin, and Koller]{TaskarB2004}
Ben Taskar, Carlos Guestrin, and Daphne Koller.
\newblock Max-margin markov networks.
\newblock In Sebastian Thrun, Lawrence Saul, and Bernhard {Sch\"{o}lkopf},
  editors, \emph{NIPS}. MIT Press, Cambridge, MA, 2004.

\bibitem[Tsochantaridis et~al.(2005)Tsochantaridis, Joachims, Hofmann, and
  Altun]{TsochantaridisI2005}
Ioannis Tsochantaridis, Thorsten Joachims, Thomas Hofmann, and Yasemin Altun.
\newblock Large margin methods for structured and interdependent output
  variables.
\newblock \emph{JMLR}, 6:\penalty0 1453--1484, 2005.
\newblock ISSN 1532-4435.

\bibitem[Besag(1975)]{BesagJ1975}
Julian Besag.
\newblock Statistical analysis of non-lattice data.
\newblock \emph{The Statistician}, 24\penalty0 (3):\penalty0 179--195, 1975.

\bibitem[Sutton and Mccallum(2005)]{SuttonC2005}
C.~Sutton and A.~Mccallum.
\newblock {Piecewise Training for Undirected Models}.
\newblock In \emph{UAI}, 2005.

\bibitem[Hazan and Urtasun(2010)]{HazanT2010}
Tamir Hazan and Raquel Urtasun.
\newblock A primal-dual message-passing algorithm for approximated large scale
  structured prediction.
\newblock In J.~Lafferty, C.~K.~I. Williams, J.~Shawe-Taylor, R.S. Zemel, and
  A.~Culotta, editors, \emph{NIPS}, pages 838--846. MIT Press, 2010.

\bibitem[Kakade et~al.(2002)Kakade, Teh, and Roweis]{KakadeS2002}
Sham Kakade, Yee~Whye Teh, and Sam~T. Roweis.
\newblock An alternate objective function for markovian fields.
\newblock In \emph{ICML}, pages 275--282, 2002.

\bibitem[Suzuki et~al.(2006)Suzuki, McDermott, and Isozaki]{SuzukiJ2006}
Jun Suzuki, Erik McDermott, and Hideki Isozaki.
\newblock Training conditional random fields with multivariate evaluation
  measures.
\newblock In \emph{ICCL-ACL}, pages 217--224, Stroudsburg, PA, USA, 2006.
  Association for Computational Linguistics.

\bibitem[Gross et~al.(2007)Gross, Russakovsky, Do, and Batzoglou]{GrossS2007}
Samuel~S. Gross, Olga Russakovsky, Chuong~B. Do, and Serafim Batzoglou.
\newblock Training conditional random fields for maximum labelwise accuracy.
\newblock In B.~Sch\"{o}lkopf, J.~Platt, and T.~Hoffman, editors, \emph{NIPS},
  pages 529--536. MIT Press, Cambridge, MA, 2007.

\bibitem[Taylor et~al.(2008)Taylor, Guiver, Robertson, and Minka]{TaylorM2008}
Michael~J. Taylor, John Guiver, Stephen Robertson, and Tom Minka.
\newblock {SoftRank:} optimizing non-smooth rank metrics.
\newblock In \emph{WSDM}, pages 77--86, 2008.

\bibitem[McAllester et~al.(2010)McAllester, Hazan, and Keshet]{McAllesterD2010}
David McAllester, Tamir Hazan, and Joseph Keshet.
\newblock Direct loss minimization for structured prediction.
\newblock In J.~Lafferty, C.~K.~I. Williams, J.~Shawe-Taylor, R.S. Zemel, and
  A.~Culotta, editors, \emph{NIPS}, pages 1594--1602. MIT Press, 2010.

\bibitem[Liu et~al.(2007)Liu, Xu, Xiong, and Li]{LETOR}
T.~Liu, J.~Xu, W.~Xiong, and H.~Li.
\newblock {LETOR}: {B}enchmark dataset for search on learning to rank for
  information retrieval.
\newblock In \emph{ACM SIGIR Workshop on Learning to Rank for Information
  Retrieval}, 2007.

\bibitem[Petterson et~al.(2009)Petterson, Caetano, McAuley, and
  Yu]{Vis_GraphMatching}
J.~Petterson, T.~S. Caetano, J.~J. McAuley, and J.~Yu.
\newblock Exponential family graph matching and ranking.
\newblock In \emph{Neural Information Processing Systems}, pages 1455--1463,
  2009.

\end{thebibliography}

\end{document}